\documentclass[10pt,twocolumn,letterpaper]{article}

\usepackage[pagenumbers]{cvpr} %

\usepackage{booktabs}
\usepackage{adjustbox}
\usepackage{multirow}

\definecolor{cvprblue}{rgb}{0.21,0.49,0.74}
\usepackage[pagebackref,breaklinks,colorlinks,allcolors=cvprblue]{hyperref}

\def\confName{CVPR}
\def\confYear{2027}

\title{LAGR: Language-Aware Grasp Re-Ranking for Task Oriented Grasping}

\newcommand{\emaillink}[2]{\href{mailto:#1}{{\normalcolor #2}}}

\author{
{
Manav Kulshrestha$^{1,\ast}$ \quad
Shreyans Jain$^{2,\ast}$ \quad
Apoorva Vashisth$^1$
}\\[2pt]
{
Ankush Gupta$^1$ \quad
Damon Conover$^3$ \quad
Aniket Bera$^1$
}\\[4pt]
$^1$Purdue University, West Lafayette, IN, USA\\
$^2$IIT Delhi, New Delhi, India\\
$^3$DEVCOM ARL, Adelphi, MD, USA\\[3pt]
{\ttfamily\small
\{%
\emaillink{mkulshre@purdue.edu}{mkulshre},%
\emaillink{vashista@purdue.edu}{vashista},%
\emaillink{gupt1681@purdue.edu}{gupt1681},%
\emaillink{aniketbera@purdue.edu}{aniketbera}%
\}@purdue.edu%
}\\
{\ttfamily\small
\emaillink{shreyansjain1208@gmail.com}{shreyansjain1208@gmail.com}
\quad
\emaillink{damon.m.conover.civ@army.mil}{damon.m.conover.civ@army.mil}
}
}

\begin{document}
\maketitle

\begingroup
\renewcommand{\thefootnote}{$\ast$}
\footnotetext{Equal Contribution.}
\endgroup

\begin{abstract}
Grasping is a fundamental capability for robotic manipulation, with the appropriate grasp depending on the downstream task.
For example, a hammer should be held by its handle for hammering but by its head when it is handed over. 
Selecting such a grasp requires consideration of two criteria: physical feasibility on the object and suitability for the task. 
Developing such an approach requires dense grasp labels with both feasibility and task scores on real observations, but existing task-oriented grasping datasets are small, synthetic, or costly to annotate. 
To address this gap, we construct a dataset of $2.68$ million scored $5$-DoF grasp annotations over $100,911$ image-task pairs from $17$ object categories. 
Additionally, we introduce LAGR, a task-oriented grasp detection method that explicitly addresses both these challenges. 
Given vision-language inputs, LAGR first generates task-independent grasp proposals along with a physical feasibility score, and then scores the generated grasps with a language-conditioned re-ranker that reasons over all proposals jointly. 
Experiments show that LAGR outperforms the next best method by $9.3$ percentage points. 
We further validate LAGR on real objects by executing task-oriented grasps with a Franka Research 3 robot.%
\end{abstract}
    
\section{Introduction}
\label{sec:intro}

Autonomous robots~\cite{vashisth2026corelin, vashisth2026scalable, kulshrestha2023structural} have the potential to transform everyday life by assisting humans in a wide range of physical tasks. 
A fundamental requirement for such autonomy is robotic manipulation~\cite{Kulshrestha2026TGrasp6DTV, Kulshrestha2025VLADGraspZG}, which enables robots to interact with objects in their surroundings. 
Central to successful manipulation is task-oriented grasping~\cite{vuong2024language}, where robots must select grasps that are not only physically stable but also appropriate for the intended task, making it a particularly challenging problem.
This challenge is further compounded by the lack of large-scale datasets that can provide task-conditioned grasp supervision. 
Existing datasets~\cite{handaliros23, depierre2018jacquard, lenz2015deep, vuong2024grasp, murali2020taskgrasp} often suffer from limited object and task diversity, sparse grasp annotations, or focus on affordance masks that do not explicitly capture grasp geometry and physical feasibility. 
Moreover, manually annotating grasps in this manner is labor-intensive, making it difficult to construct comprehensive datasets for learning task-oriented grasping.

\begin{figure}[t]
    \centering
    \includegraphics[
        width=\linewidth,
        trim={2.8cm 4cm 3.5cm 1.3cm},
        clip
    ]{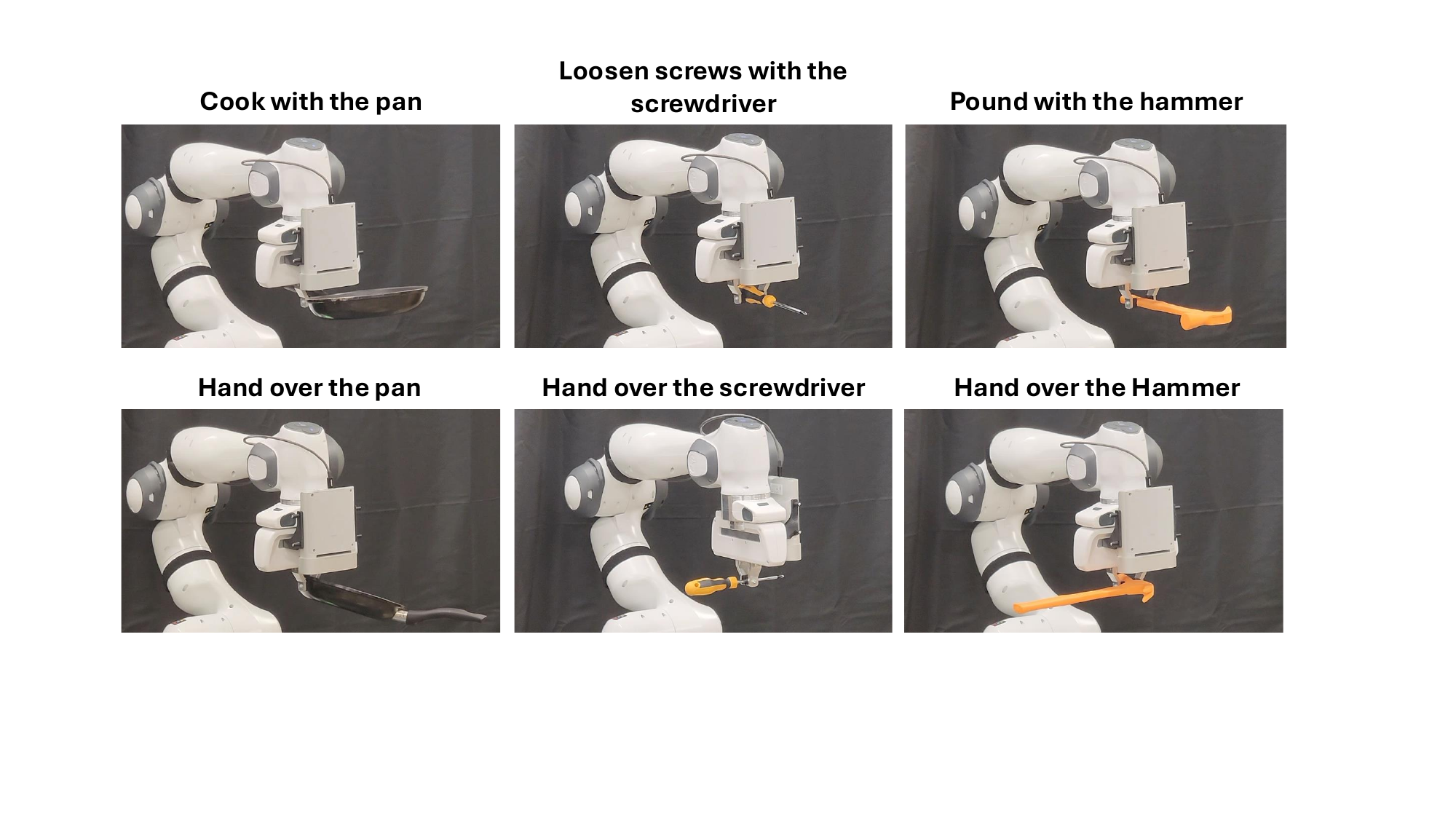}
    \caption{Examples of LAGR generating distinct task-specific grasps for the same real-world objects. Functional tasks favor grasps that enable effective tool use, whereas handover tasks favor grasps that leave the handles accessible to the recipient.}
    \label{fig:teaser}
\end{figure}

Existing grasp detection methods~\cite{mahler2017dexnet, morrison2018closing, fang2020graspnet, eppner2021acronym, yu2022se} focus on physical grasp quality without explicitly considering task intent. 
Task-oriented and language-guided approaches~\cite{sun2021gater, holomjova2023one, tziafas2023language, vuong2024language, li2024shapegrasp} incorporate semantic and functional reasoning to identify task-appropriate grasps. 
However, these methods do not consistently distinguish task-independent physical feasibility from task-dependent grasp suitability, limiting their ability to explicitly evaluate and compare feasible grasp configurations under different manipulation goals.
While affordance learning methods~\cite{wu2025ragnet, ma2025glover++, zhang2026a4} focus on identifying task-relevant object regions, they do not consider physical feasibility of grasp configurations. 
Existing grasp datasets~\cite{handaliros23, depierre2018jacquard, lenz2015deep, vuong2024grasp} provide extensive geometric supervision but largely lack task annotations, while task-oriented datasets~\cite{murali2020taskgrasp, tang2023graspgpt, tang2025foundationgrasp} often suffer from limited object–task diversity and sparse grasp-level supervision. 
Furthermore, partial observations introduce uncertainty in 3D 6-DoF grasp estimation, motivating image-space 5-DoF representations grounded in visible geometry. 
These limitations highlight the need for large-scale, language-conditioned grasp annotations that jointly capture geometric feasibility and functional suitability.

To address this limitation, we introduce a large-scale dataset of 2.68 million scored grasps across 100,911 image–task pairs. 
Additionally, we build upon this dataset and propose LAGR, a multimodal grasp generation framework that explicitly decouples physical feasibility from task-dependent grasp selection. 
LAGR first generates a shared set of physically plausible grasp proposals from visual observations, then employs a language-conditioned reranker that jointly reasons over proposal geometry, visual context, and task intent to select the most appropriate grasp. 
Our main contributions are summarized as follows:
\begin{itemize}

    \item Our dataset comprises $2.68$ million scored grasp annotations across $100,911$ image–task pairs, with each annotation providing explicit grasp geometry, a physical feasibility score, and a task-compatibility score conditioned on the corresponding natural-language instruction.

    \item We propose LAGR, a multimodal architecture that separates task-independent physical grasp proposal generation from language-conditioned selection, leveraging joint reasoning over candidate grasps to identify functionally appropriate grasp configurations.

    \item We demonstrate that LAGR achieves $85.9\%$ task-conditioned grasp accuracy on our dataset, outperforming the strongest baseline by $9.3$ percentage points. 
\end{itemize}
We also validate LAGR's performance through real-world task-oriented grasping via a Franka Research 3 robot.

\section{Related Work}
\label{sec:related}

\textbf{Grasping and Affordance datasets}.
Existing grasping datasets can be broadly categorized based on their supervision and intended applications. 
Task-agnostic datasets, including Cornell \cite{lenz2015deep}, Jacquard \cite{depierre2018jacquard}, Jacquard V2 \cite{li2024jacquard}, and OCID-Grasp \cite{ainetter2021end}, provide explicit grasp annotations but do not account for downstream manipulation goals, potentially favoring physically feasible yet functionally unsuitable grasps. 
Language-driven datasets, such as Grasp-Anything \cite{vuong2024grasp} and Grasp-Anything++ \cite{vuong2024language}, substantially expand object diversity and language-guided grasp supervision through synthetic data, but do not explicitly distinguish physical feasibility from task-dependent grasp suitability in real-world observations. 
Task-oriented 6-DoF grasping datasets, including TaskGrasp \cite{murali2020taskgrasp}, LA-TaskGrasp \cite{tang2023graspgpt}, and LaViA-TaskGrasp \cite{tang2025foundationgrasp}, incorporate object–task relationships and semantic knowledge, but lack dense, image-aligned grasp annotations with independently quantified physical feasibility and task-dependent utility. 
Finally, affordance datasets, such as RAGNet \cite{wu2025ragnet}, HOVA-500K \cite{ma2025glover++}, and HANDAL \cite{handaliros23}, provide task-relevant interaction regions, affordance annotations, or reconstructed object geometry without directly supervising explicit task-conditioned grasp configurations. 
To bridge these gaps, we introduce a geometry-grounded dataset comprising 2.68 million scored grasps across 100,911 real-world image–task pairs, explicitly associating grasp geometry with physical feasibility and task-dependent suitability.

\noindent\textbf{Task-oriented Grasping}.
Existing task-oriented grasping methods can be broadly categorized into three groups based on how they incorporate task information. 
Semantic reasoning-based approaches~\cite{sun2021gater, holomjova2023one, li2024shapegrasp} leverage object–action relationships, affordance transfer, or geometric decomposition to identify functionally suitable grasp regions. 
However, these methods primarily emphasize semantic part selection and do not explicitly optimize grasp-level physical feasibility. %
Language-conditioned grasp generation methods~\cite{tziafas2023language, vuong2024language} integrate visual and linguistic representations to directly predict grasp configurations. 
While effective at language-guided object localization and grasp synthesis, they do not explicitly separate task-independent physical grasp generation from task-dependent selection among feasible candidates. 
This can cause the grasp to be semantically appropriate for a task without being physically feasible to execute and vice-versa.
Affordance reasoning approaches~\cite{wu2025ragnet, ma2025glover++, zhang2026a4} employ vision-language models, human demonstrations, and semantic reasoning to identify task-relevant interaction regions. 
However, region-level affordance predictions do not directly specify executable grasp configurations or quantify their geometric feasibility. 
In contrast, LAGR explicitly decouples physical grasp generation from task-conditioned selection, enabling joint reasoning over multiple geometrically grounded grasp proposals to identify configurations that satisfy both physical and functional needs.

\section{Dataset}
\label{sec:dataset}

\begin{figure}
    \centering
    \includegraphics[width=\linewidth]{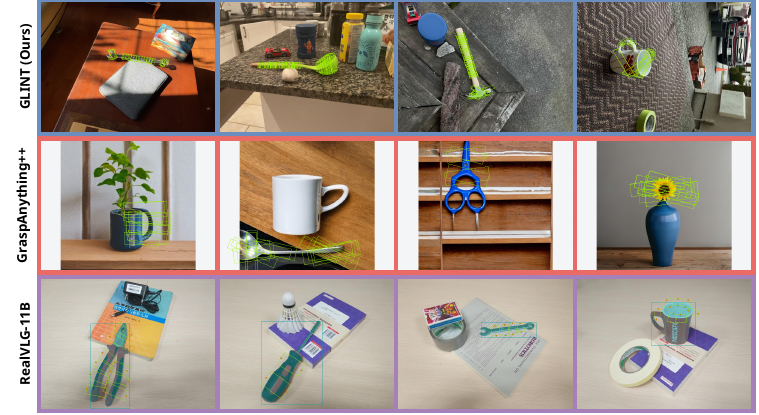}
    \caption{Comparison of representative grasp annotations from existing datasets and LAGR. }
    \label{fig:dataset_comparison}
    \vspace{-5mm}
\end{figure}

\textbf{Overview.}
Our dataset consists of observation–instruction pairs, each labeled with multiple grasp annotations. Furthermore, each grasp annotations is labeled with two scores: a \textbf{physical feasibility score} and a \textbf{task-conditioned annotation score}.
Each observation comprises of an RGB image, an aligned depth map, a semantic segmentation mask of the scene, and a handle-part mask identifying the object's functional region for grasping. 
For our initial dataset construction, we leverage the HANDAL dataset~\cite{handaliros23}, which provides RGB images, aligned NeRF-derived depth maps, camera intrinsics, object poses, reconstructed 3D meshes, visible-object masks, and handle-part masks. 
We use these available annotations and geometric representations to generate our observation–instruction pairs and their associated scores.

\noindent\textbf{Problem formulation}.
For each observation–instruction pair $(\mathcal{V}, q)$, where $\mathcal{V}$ denotes the multimodal visual observations and $q$ represents a natural-language instruction specifying the task, our dataset provides a set of $m \in \mathbb{N}^{+}$ valid grasp candidates, denoted by $\mathcal{G} = \{g_k\}_{k=1}^{m}$ and their corresponding scores. 
Each grasp candidate $g_k$ is associated with two scores: a geometry-derived feasibility score $f_k \in [0,1]$ and a task-conditioned annotation score $a_k \in [0,1]$.

\noindent\textbf{Grasp generation}.
We generate candidate grasps $\widetilde{\mathcal{G}}=\{g_k\}_{k=1}^{j}$ where $j\in \mathcal{N}^+$ by sampling locations along the visible functional regions of the object and filtering for geometric validity. 
At each sampled location, the gripper closing direction is determined from the object's orientation, while the grasp center is adjusted via opposing object boundaries. 
The estimated local thickness and gripper's physical constraints determine the required contact depth and gripper opening.
A geometrically valid grasp must establish opposing contacts with sufficient surface support, satisfy the gripper's opening constraint, and maintain adequate visibility and depth consistency between the reconstructed mesh and the observed depth map. 
We additionally reject candidates whose projected finger plates or approach regions exhibit excessive overlap with the object.

\noindent\textbf{Physical feasibility score}. 
Following geometric validity filtering, we evaluate each remaining grasp candidate $g_k \in \tilde{\mathcal{G}}$ using a set of normalized quality metrics that assess visibility, depth consistency, gripper opening, orientation reliability, object support, contact symmetry, and collision clearance. 
These metrics are aggregated into a physical feasibility score $f_k \in [0,1]$ using a weighted geometric mean:
\[
f_k = S_{\mathrm{feas}}(g_k)
= \exp\left(
\frac{\sum_{i=1}^{I}\lambda_i
\log\left(\max(s_i(g_k),\epsilon)\right)}
{\sum_{i=1}^{I}\lambda_i}
\right),
\]
where $s_i(g_k)\in[0,1]$ denotes the $i$-th quality metric, $\lambda_i>0$ is its corresponding weight, $I\in\mathbb{N}^+$ is the number of metrics, and $\epsilon=10^{-4}$ ensures numerical stability. 
We assign higher weights to contact support and collision clearance, as these directly reflect the grasp's contact reliability and potential for collision. 
The geometric mean penalizes low-scoring components, preventing strong performance in other metrics from fully compensating for critical geometric weaknesses.
We retain only candidates whose feasibility scores exceed a predefined threshold $\tau_f=0.52$, yielding the physically feasible grasp set $\mathcal{G}_{\mathrm{feas}}=\{g_j\in\mathcal{G}\;|\;f_j\geq\tau_f\}$.
The resulting $m=|\mathcal{G}_{\mathrm{feas}}|$ grasp candidates, together with their corresponding feasibility scores, are passed to the subsequent task-conditioned annotation stage. 

\noindent\textbf{Task-specific utility score}.
Given the feasibility-filtered grasp set $\mathcal{G}_{\mathrm{feas}}$, we assign each candidate $g_k\in\mathcal{G}_{\mathrm{feas}}$ a task-specific utility score $u(g_k,q)\in[0,1]$, reflecting its suitability for the instruction $q$. 
We consider three task categories: object transport, functional use, and human handover.
For object transport, all retained grasps are assigned equal task utility, irrespective of their location on the object $u(g_k,q=\mathrm{transport})=1$.
For functional use, we prioritize grasps on the object's handle or designated graspable region, with preferred locations determined by category-specific mechanical considerations, including power, precision, balance, load control, and ergonomics. 
For handover, we prioritize grasps on the functional region to preserve receiver's handle accessibility. 
Let \(\mathcal{G}_{F}(q)\subseteq\mathcal{G}_{\mathrm{feas}}\) denote the set of functional-region grasps retained for task $q$. 
When \(\mathcal{G}_{F}(q)=\varnothing\), we consider handle grasps as a fallback, assigning higher utility to configurations leaving a greater portion of the handle accessible.

\noindent\textbf{Task-conditioned Annotation Score}.
Finally, we combine physical feasibility with task utility to obtain the task-conditioned annotation score $a_k=f_k\times u(g_k,q)$.
This ensures that the final annotation jointly reflects geometric grasp quality and task suitability, while restricting the annotated candidates to those that satisfying feasibility constraints.

\section{Method}
\label{sec:method}

\begin{figure*}[htbp]
    \centering
    \includegraphics[width=\linewidth]{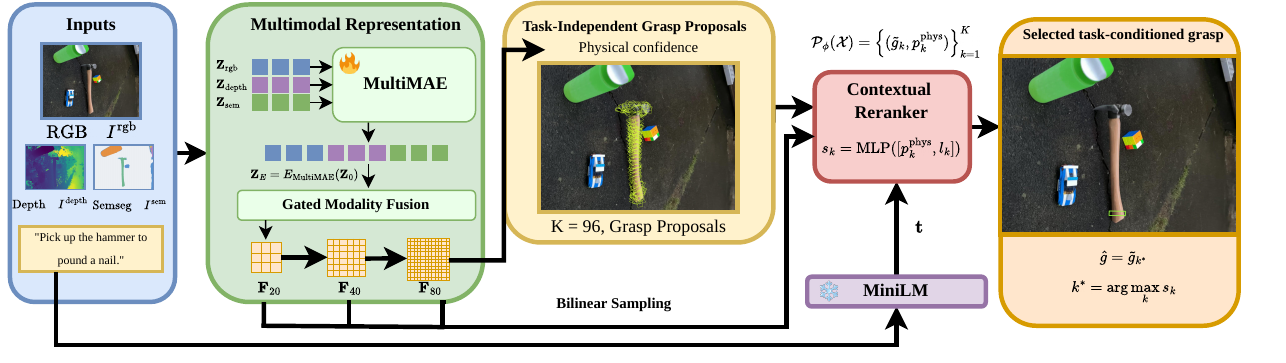}
    \caption{Overview of LAGR. Given multimodal visual observations, a task-independent proposal generator extracts fused visual features and predicts physically plausible grasp candidates with associated confidence scores. A language-conditioned re-ranker combines proposal geometry, multi-scale visual features, and task instruction embeddings, then applies a proposal-context transformer to jointly evaluate candidate suitability. The final grasp is selected by combining task-compatibility predictions with physical confidence, without modifying the proposed grasp geometry.}
    \label{fig:overview_approach}
\end{figure*}

\noindent\textbf{Overview}. 
Leveraging our geometry-grounded, task-conditioned grasp dataset, we introduce LAGR, a two-stage task-oriented grasp generation framework that decouples task-independent physical grasp generation from language-conditioned grasp selection.
We adopt a two-stage architecture that explicitly separates physical grasp feasibility from task suitability. 
First, a task-independent grasp generator processes the multimodal observations using a pretrained visual encoder to generate a diverse set of physically plausible grasp candidates. 
Each grasp candidate is associated with a physical feasibility score, indicating how geometrically plausible and accessible a candidate grasp is. 
Next, a language-conditioned grasp re-ranker evaluates these candidates based on the natural-language task instruction, their geometric properties, and the corresponding visual features. 
Instead of evaluating each grasp independently, our re-ranker jointly reasons over all candidate grasps to determine their relative suitability for the intended manipulation task. 
Finally, the task-compatibility scores are combined with the physical feasibility scores to select a grasp that is both geometrically plausible and functionally appropriate. 

\noindent\textbf{Problem Formulation}. 
Given a multimodal observation $\mathcal{V}=(I^{\mathrm{rgb}},I^{\mathrm{depth}},I^{\mathrm{sem}})$ and a natural-language instruction $q$ specifying the intended manipulation task, LAGR predicts $n\in \mathbb{N}^+$ planar, parallel-jaw grasp candidates $\{g_k\}_{k=1}^n$ and their corresponding task-conditioned suitability scores $F_{\Theta}(\mathcal{V},q) = \{(g_k,s_k)\}_{k=1}^{n},$.
We represent each grasp as an oriented rectangle parameterized by $g_k=(c_{y,k},c_{x,k},\theta_k,w_k,h_k)$, where $(c_{y,k},c_{x,k})$ denotes the grasp center in image coordinates, $\theta_k$ specifies the in-plane orientation of the gripper closing axis, $w_k$ represents the gripper opening, and $h_k$ denotes the fixed jaw-contact depth. 
We adopt a 5-DoF image-space grasp representation instead of 3D 6-DoF grasp estimation, as single-view observations provide incomplete 3D geometry due to occlusions, making image-space prediction more directly grounded in the available visual information.
The grasp orientation is uniquely defined within the interval $[0,\pi)$ as parallel-jaw grasps are invariant under a rotation of $\pi$.
Accordingly, we measure the angular discrepancy between two grasps $g_1$ and $g_2$ as $\Delta_{\pi}(\theta_1,\theta_2)=\arccos\!\left(\left|\cos(\theta_1-\theta_2)\right|\right)$.

\noindent\textbf{Task-independent proposal generator}. 
Given a multimodal visual observation \(\mathcal{V}\), our grasp generator \(\mathcal{P}_{\phi}\) predicts a fixed-size set of \(n\) grasp proposals and their corresponding physical confidence scores $\mathcal{P}_{\phi}(\mathcal{V})=\{(g_k,p_k)\}_{k=1}^{n}$.

We employ a MultiMAE~\cite{bachmann2022multimae} encoder to extract joint visual representations from RGB images, depth maps, and semantic segmentation masks. 
Modality-specific patch embeddings are jointly processed by the transformer, allowing cross-modal interactions before feature extraction. 
The resulting modality-associated token representations are reshaped into spatial feature maps and adaptively fused using learned modality gates. 
A convolutional fusion module produces a shared feature representation, which is progressively upsampled through a multi-scale feature neck. 
The highest-resolution feature map is used for dense grasp prediction, while all three resolutions are retained for subsequent proposal-level feature extraction.
A convolutional prediction head estimates grasp parameters over a spatial grid with discrete orientation bins. 
For each spatial location and orientation bin, the head predicts a grasp-center heatmap, physical feasibility, subpixel center offsets, an angular residual, gripper opening, and jaw depth. 

We combine the predicted center likelihood and physical feasibility to obtain the physical confidence of each dense hypothesis. 
This confidence serves as a task-independent measure of grasp plausibility and is used for proposal selection and subsequent task-conditioned ranking.

The dense head produces the grasp hypotheses. 
We first identify spatial local maxima of the physical-confidence map within each orientation bin and select the top hypotheses across all bins. 
Each selected hypothesis is decoded into an oriented grasp rectangle using its predicted center offsets, orientation bin and angular residual, gripper opening, and jaw depth.
To promote diversity, we apply greedy suppression based on the spatial and angular proximity of decoded grasps -- a candidate is suppressed when both its center distance and angular difference from a previously selected proposal fall below predefined thresholds. 
The resulting proposals are ranked by physical confidence, and up to $n$ candidates are retained. 
When fewer than $n$ candidates remain, the proposal tensor is padded to maintain a fixed size. 
The resulting grasp proposals and their physical confidence scores are subsequently passed to the task-conditioned reranking module. 

\noindent\textbf{Grasp re-ranker}. 
Given the instruction-independent grasp proposals \(\mathcal{P}_{\phi}(\mathcal{V})=\{g_k,p_k)\}_{k=1}^{n}\) and a natural-language instruction \(q\), the task-conditioned reranker predicts the suitability score of each proposal for the specified task.

For each candidate \(g_k\), we bilinearly sample features at its grasp center from the multi-scale visual representations. 
To incorporate grasp geometry, we separately encode the normalized grasp center, orientation, gripper opening, jaw depth, and physical confidence. 
The orientation is represented by \((\cos 2\theta_k,\sin 2\theta_k)\) to account for the \(\pi\)-periodicity.

We encode the task instruction using a frozen MiniLM language encoder~\cite{wang2020minilm}, followed by a learned projection to obtain \(\mathbf{t}_q\in\mathbb{R}^{256}\). 
To capture interactions between grasp characteristics and task semantics, we construct a fused representation using concatenation, element-wise multiplication, and subtraction $\mathbf{z}_k=\psi\left(\left[\mathbf{p}_k;\,\mathbf{t}_q;\,\mathbf{p}_k\odot\mathbf{t}_q;\,\mathbf{p}_k-\mathbf{t}_q\right]\right)+\mathbf{p}_k$,
where $\psi$ denotes a learned fusion network, $\odot$ represents element-wise multiplication, and $[\cdot;\cdot]$ denotes concatenation. 
$\mathbf{p}_k = \mathbf{v}_k+ \Phi_{\mathrm{geom}}(\mathbf{r}_k)$ where $\mathbf{v}_k$ is the visual feature vector that is extracted at the center of the $k^{\text{th}}$ grasp proposal and $\Phi_{\mathrm{geom}}(\mathbf{r}_k)$ is the learned geometry projection specific to geometry vector $\mathbf{r}_k$ of $k^{\text{th}}$ grasp.
The fused proposal embeddings are subsequently processed by a proposal-context transformer, enabling each grasp candidate to be evaluated in relation to the other candidates within the same observation.

A prediction head maps each contextualized proposal embedding to a task-compatibility logit $l_k$. 
The final grasp selection combines the predicted task compatibility with the physical confidence $p_k$, jointly accounting for grasp plausibility and task relevance. 
A two-layer MLP with a 16-dimensional hidden layer and ReLU activation maps the concatenated inputs $[p_k,l_k]$ to the scalar ranking score $S_{\mathrm{rank}}(g_k,q)$. %
Importantly, the reranker operates exclusively on the generated candidates and does not modify their grasp geometry. 
The highest-ranked candidate is selected from the original proposal set $k^*=\arg\max_{k\in\{1,\ldots,n\}}S_{\mathrm{rank}}(g_k,q)$ such that $g^*= g_{k^*}$, where \(S_{\mathrm{rank}}\) denotes the combined ranking score. 

\noindent\textbf{Loss Function}. 
The grasp proposal generator is trained using a task-independent physical loss that combines a CenterNet-style heatmap objective ~\cite{zhou2019objects}, geometric regression, Gaussian Wasserstein alignment adapted from rotated object detection ~\cite{yang2022rethinkingrotatedobjectdetection}, and feasibility supervision, $\mathcal{L}_{\mathrm{phys}}=\mathcal{L}_{\mathrm{heat}}+2(\mathcal{L}_{\mathrm{offset}}+\mathcal{L}_{\mathrm{angle}}+\mathcal{L}_{\mathrm{size}})+2\mathcal{L}_{\mathrm{GWD}}+0.5\mathcal{L}_{\mathrm{quality}}$. 
Here, the heatmap loss supervises grasp localization, regression losses refine grasp geometry, Gaussian Wasserstein loss improves rectangular alignment, and the quality loss supervises geometric feasibility. 
This objective enables the generator to learn physically plausible grasps independently of task semantics.
The grasp reranker is trained using a task-conditioned ranking loss,
\begin{equation}
\mathcal{L}_{\mathrm{rank}}=\mathcal{L}_{\mathrm{cal}}+\mathcal{L}_{\mathrm{list}}
\label{eq:loss}
\end{equation}
Here, the calibration loss $\mathcal{L}_{\mathrm{cal}}$ aligns predicted task-compatibility scores with annotation-derived utility targets, while the listwise loss $\mathcal{L}_{\mathrm{list}}$ encourages higher-utility grasps to rank above less suitable candidates.

\section{Experiments}
\label{sec:expts}

\subsection{Setup}
\textbf{Benchmark}.
We evaluate LAGR on our task-oriented grasping benchmark derived from HANDAL \cite{handaliros23}, comprising 2.68 million scored grasp annotations across 100,911 image–task pairs, spanning 17 object categories and 51 category–task combinations. 
The grasp generator predicts $96$ candidate grasps per observation, which are subsequently evaluated by the language-conditioned reranker.

\begin{table*}[t]
\centering
\small
\setlength{\tabcolsep}{7.5pt}
\renewcommand{\arraystretch}{1.12}
\begin{adjustbox}{max width=\textwidth}
\begin{tabular}{@{}lccccccccc@{}}
\toprule

\multirow{2}{*}{\textbf{Approach}}
    & \multicolumn{3}{c}{\textbf{Task-Conditioned}}
    & \multicolumn{3}{c}{\textbf{Grasp Stability}}
    & \multirow{2}{*}{\shortstack{\textbf{GCR}\\(\%) $\downarrow$}}
    & \multirow{2}{*}{\shortstack{\textbf{No Output}\\(n / 1650) $\downarrow$}}
    & \multirow{2}{*}{\shortstack{\textbf{Time}\\(ms / sample) $\downarrow$}} \\

\cmidrule(lr){2-4}
\cmidrule(lr){5-7}

    & \shortstack{\textbf{SR}\\(\%) $\uparrow$}
    & \textbf{IoU} $\uparrow$
    & \shortstack{\textbf{Angle Err.}\\($^\circ$) $\downarrow$}
    & \shortstack{\textbf{SR}\\(\%) $\uparrow$}
    & \textbf{IoU} $\uparrow$
    & \shortstack{\textbf{Angle Err.}\\($^\circ$) $\downarrow$}
    &
    &
    & \\

\midrule

OS-TOG~\cite{holomjova2023one}
    & 44.8
    & $0.50$ {\footnotesize $\pm 0.31$}
    & \underline{$8.48$ {\footnotesize $\pm 14.3$}}
    & 54.0
    & \underline{$0.60$ {\footnotesize $\pm 0.24$}}
    & \underline{$6.96$ {\footnotesize $\pm 10.97$}}
    & \underline{$1.59$ {\footnotesize $\pm 8.87$}}
    & 632
    & 3789 \\

SE-ResUNet~\cite{yu2022se}
    & 59.6
    & $0.35$ {\footnotesize $\pm 0.28$}
    & $9.81$ {\footnotesize $\pm 15.1$}
    & 75.7
    & $0.45$ {\footnotesize $\pm 0.24$}
    & $9.46$ {\footnotesize $\pm 14.7$}
    & $9.26$ {\footnotesize $\pm 22.13$}
    & \underline{3}
    & \textbf{9} \\

GATER~\cite{sun2021gater}
    & \underline{76.6}
    & \underline{$0.51$ {\footnotesize $\pm 0.28$}}
    & $9.64$ {\footnotesize $\pm 12.7$}
    & \underline{80.9}
    & $0.54$ {\footnotesize $\pm 0.26$}
    & $9.62$ {\footnotesize $\pm 12.8$}
    & $1.93$ {\footnotesize $\pm 8.80$}
    & \textbf{0}
    & \underline{23.18} \\

LGD$^\dagger$~\cite{vuong2024language}
    & 1.45
    & $0.03$ {\footnotesize $\pm 0.08$}
    & $35.8$ {\footnotesize $\pm 25.3$}
    & 2.36
    & $0.05$ {\footnotesize $\pm 0.09$}
    & $35.9$ {\footnotesize $\pm 25.1$}
    & $7.24$ {\footnotesize $\pm 16.5$}
    & \textbf{0}
    & 2633 \\

LGD~\cite{vuong2024language}
    & 5.94
    & $0.05$ {\footnotesize $\pm 0.12$}
    & $33.7$ {\footnotesize $\pm 25.1$}
    & 9.4
    & $0.08$ {\footnotesize $\pm 0.14$}
    & $34.2$ {\footnotesize $\pm 25.3$}
    & $24.9$ {\footnotesize $\pm 36.7$}
    & \textbf{0}
    & 2607 \\

RealVLG-R1~\cite{li2026realvlg}
    & $10.2$
    & $0.11$ {\footnotesize $\pm 0.13$}
    & $39.4$ {\footnotesize $\pm 25.2$}
    & $16.3$
    & $0.18$ {\footnotesize $\pm 0.13$}
    & $39.9$ {\footnotesize $\pm 24.9$}
    & $54.8$ {\footnotesize $\pm 34.0$}
    & \textbf{0}
    & 4133 \\

RAGNet~\cite{wu2025ragnet}
    & 25.6
    & $0.15$ {\footnotesize $\pm 0.13$}
    & $14.9$ {\footnotesize $\pm 19.6$}
    & 39.7
    & $0.24$ {\footnotesize $\pm 0.09$}
    & $14.1$ {\footnotesize $\pm 18.4$}
    & $2.93$ {\footnotesize $\pm 10.6$}
    & \textbf{0}
    & 1915 \\

CROG~\cite{tziafas2023language}
    & $23.1$
    & $0.26$ {\footnotesize $\pm 0.17$}
    & $44.6$ {\footnotesize $\pm 25.9$}
    & $24.5$
    & $0.28$ {\footnotesize $\pm 0.16$}
    & $44.3$ {\footnotesize $\pm 26.0$}
    & $57.7$ {\footnotesize $\pm 38.6$}
    & \textbf{0}
    & 128 \\

GLOVER++~\cite{ma2025glover++}
    & 33.7
    & $0.18$ {\footnotesize $\pm 0.17$}
    & $16.0$ {\footnotesize $\pm 18.5$}
    & $49.0$
    & $0.27$ {\footnotesize $\pm 0.14$}
    & $8.13$ {\footnotesize $\pm 13.2$}
    & $16.0$ {\footnotesize $\pm 17.9$}
    & 5
    & 48 \\

A4-Agent$^\dagger$~\cite{zhang2026a4}
    & 21.5
    & $0.13$ {\footnotesize $\pm 0.15$}
    & $12.1$ {\footnotesize $\pm 17.4$}
    & $39.2$
    & $0.23$ {\footnotesize $\pm 0.14$}
    & $11.9$ {\footnotesize $\pm 17.0$}
    & $35.6$ {\footnotesize $\pm 42.7$}
    & 89
    & 18711 \\

ShapeGrasp$^\dagger$~\cite{li2024shapegrasp}
    & 10.6
    & $0.10$ {\footnotesize $\pm 0.13$}
    & $16.6$ {\footnotesize $\pm 21.9$}
    & $19.6$
    & $0.16$ {\footnotesize $\pm 0.13$}
    & $15.0$ {\footnotesize $\pm 19.6$}
    & $24.9$ {\footnotesize $\pm 24.9$}
    & 419
    & 31682 \\

\midrule

Ours
    & $\mathbf{85.9}$
    & $\mathbf{0.58}$ {\footnotesize $\mathbf{\pm 0.24}$}
    & $\mathbf{5.81}$ {\footnotesize $\mathbf{\pm 9.42}$}
    & $\mathbf{89.6}$
    & $\mathbf{0.61}$ {\footnotesize $\mathbf{\pm 0.22}$}
    & $\mathbf{5.73}$ {\footnotesize $\mathbf{\pm 9.28}$}
    & $\mathbf{0.84}$ {\footnotesize $\mathbf{\pm 5.18}$}
    & \textbf{0}
    & 37 \\

\bottomrule
\end{tabular}
\end{adjustbox}

\caption{
Quantitative comparison of task-conditioned grasp prediction and grasp stability.
SR denotes grasp success rate, IoU denotes intersection-over-union with
ground-truth grasp rectangles, and Angle Err. is grasp orientation error.
GCR denotes grasp collision rate. $\dagger$ indicates the method was
evaluated zero-shot.
}
\label{tab:main_results}
\end{table*}

\noindent\textbf{Training}.
We train on NVIDIA L40S GPUs for 30 epochs, comprising five proposal-only epochs, five reranker-only epochs, and 20 joint-training epochs. 
We use AdamW optimizer with $\beta_1=0.9$, $\beta_2=0.999$, and weight decay $10^{-4}$. 
The proposal and reranker learning rates are both $2\times10^{-4}$. 

\noindent\textbf{Real-robot Experiments}.
We evaluate LAGR in a simplified tabletop setting adapted from HANDAL~\cite{handaliros23}, where objects are placed on a table under varying orientations and manipulation instructions. 
We use a Franka Research 3 arm equipped with a two-finger parallel-jaw gripper for grasp execution. 
For each object, we consider two distinct manipulation tasks and ten object orientations, resulting in 20 image–task pairs per object. 
This controlled setup enables systematic evaluation of grasp prediction and execution across variations in object orientation and task requirements.
We report the grasp success rate (SR) as the primary metric.

\subsection{Evaluation Metrics}
We consider both task-oriented and task-agnostic grasping settings using three standard metrics—Success Rate (SR), Grasp Intersection-over-Union (IoU), and Grasp Angle Error (Angle Err.)—along with a newly introduced metric, Grasp Collision Rate (GCR), which quantifies potential gripper–object collisions in image space, an aspect not explicitly captured by the conventional evaluation metrics.

\noindent\textbf{Success Rate (SR)} measures the percentage of predicted grasps that satisfy both an IoU threshold of $0.25$ and an angular error threshold of $30^\circ$ with at least one common ground-truth grasp annotation. 
A higher SR indicates better grasp prediction performance.

\noindent\textbf{Grasp Intersection-over-Union (IoU)} measures the spatial overlap between the predicted grasp rectangle and the ground-truth grasp annotations, taking the maximum overlap across all valid annotations. 
A higher IoU indicates better geometric alignment, with $1$ representing perfect overlap and $0$ indicating no overlap.

\noindent\textbf{Grasp Angle Error (Angle Err.)} measures the angular difference between the predicted grasp orientation and the ground-truth grasp with the highest IoU, accounting for the rotational symmetry of a parallel-jaw gripper. Samples with no IoU overlap with any ground-truth grasp are excluded from this metric. Lower values indicate better orientation alignment, with $0^\circ$ representing parallel and $90^\circ$ representing perpendicular orientations.

\noindent\textbf{Grasp Collision Rate (GCR)} quantifies potential gripper–object interference, which conventional grasp evaluation metrics do not explicitly capture. Given an object mask $M_{\mathcal V}$ and the two opposing gripper-finger edges $L_{1,g^*}$ and $L_{2,g^*}$ of the predicted grasp $g^*$, we define
\[
\mathrm{GCR}(g^*)=\frac{|L_{1,g^*}\cap M_{\mathcal V}|+|L_{2,g^*}\cap M_{\mathcal V}|}{|L_{1,g^*}|+|L_{2,g^*}|}.
\]
GCR measures the fraction of the projected finger edges overlapping the object mask. Lower values indicate less potential interference, with $0$ representing no overlap and $1$ indicating complete overlap.

\begin{table*}[t]
\centering
\scriptsize
\setlength{\tabcolsep}{1pt}
\renewcommand{\arraystretch}{1.12}
\begin{tabular*}{\textwidth}{@{\extracolsep{\fill}}lcccccccc@{}}
\toprule
\textbf{Configuration}
& \multicolumn{3}{c}{\textbf{Task-conditioned}}
& \multicolumn{3}{c}{\textbf{Any-task GT}}
& \shortstack{\textbf{GCR}}
& \shortstack{\textbf{Time}} \\

\cmidrule(lr){2-4}\cmidrule(lr){5-7}
 & \textbf{SR (\%)}
& \textbf{IoU}
& \textbf{Angle ($^\circ$)}
& \textbf{SR (\%)}
& \textbf{IoU}
& \textbf{Angle ($^\circ$)}
& $\%$ & ms/sample \\

\midrule

\midrule

Architectural Components & & & & & \\

\ \ \ \ Physical-confidence selection only 
& 53.03 & $0.370 \pm 0.352$ & $3.23 \pm 7.08$ & 88.00 & $0.619 \pm 0.246$ & $2.97 \pm 5.98$ & $0.90 \pm 6.06$ & 38.95 \\

\ \ \ \ Without proposal geometry
& 74.00 & $0.488 \pm 0.295$ & $7.92 \pm 12.97$ & 79.09 & $0.523 \pm 0.276$ & $8.10 \pm 13.40$ & $1.90 \pm 8.56$ & 38.97 \\

\ \ \ \ Without proposal-context transformer
& 77.76 & $0.515 \pm 0.275$ & $9.06 \pm 13.10$ & 81.27 & $0.540 \pm 0.260$ & $9.09 \pm 13.06$ & $1.85 \pm 8.64$ & 41.48 \\

\ \ \ \ Without physical-quality score
& 80.79 & $0.538 \pm 0.263$ & $8.55 \pm 11.93$ & 84.24 & $0.561 \pm 0.246$ & $8.58 \pm 11.92$ & $1.90 \pm 9.04$ & 40.51 \\

\midrule

Reranker Loss Components & & & & & \\

\ \ \ \ Calibration only
& 77.52 & $0.530 \pm 0.292$ & $5.77 \pm 10.86$ & 86.85 & $0.597 \pm 0.244$ & $5.75 \pm 10.97$ & $1.70 \pm 8.49$ & 42.17 \\

\ \ \ \ Calibration + margin
& 77.94 & $0.533 \pm 0.288$ & $6.09 \pm 11.19$ & 85.76 & $0.589 \pm 0.249$ & $6.08 \pm 11.32$ & $1.62 \pm 8.12$ & 44.88 \\

\ \ \ \ Calibration + listwise (Ours)
& 82.24 & $0.551 \pm 0.263$ & $7.46 \pm 11.91$ & 86.06 & $0.577 \pm 0.241$ & $7.48 \pm 11.85$ & $1.69 \pm 8.07$ & 43.40 \\

\midrule 

Number of Grasp Proposals $n$ & & & & & \\

\ \ \ \ 1
& 53.03 & $0.370 \pm 0.352$ & $3.23 \pm 7.08$ & 88.00 & $0.619 \pm 0.246$ & $2.97 \pm 5.98$ & $0.90 \pm 6.06$ & 31.71 \\

\ \ \ \ 8 
& 76.67 & $0.521 \pm 0.295$ & $5.95 \pm 10.11$ & 82.79 & $0.566 \pm 0.268$ & $5.88 \pm 10.13$ & $1.41 \pm 7.51$ & 32.77 \\

\ \ \ \ 16 
& 78.97 & $0.532 \pm 0.282$ & $6.87 \pm 10.65$ & 83.03 & $0.561 \pm 0.262$ & $6.91 \pm 10.81$ & $1.60 \pm 7.89$ & 31.94 \\

\ \ \ \ 32 
& 80.85 & $0.544 \pm 0.271$ & $7.27 \pm 10.87$ & 84.18 & $0.567 \pm 0.254$ & $7.41 \pm 11.19$ & $1.53 \pm 7.61$ & 33.00 \\

\ \ \ \ 64 
& 82.36 & $0.553 \pm 0.263$ & $7.11 \pm 10.36$ & 85.64 & $0.576 \pm 0.245$ & $7.30 \pm 10.72$ & $1.59 \pm 7.92$ & 35.38 \\

\ \ \ \ 96 
& 82.06 & $0.552 \pm 0.263$ & $7.27 \pm 10.56$ & 85.33 & $0.575 \pm 0.246$ & $7.36 \pm 10.69$ & $1.75 \pm 8.39$ & 37.77 \\

\ \ \ \ 128 
& 82.61 & $0.553 \pm 0.261$ & $7.28 \pm 10.70$ & 86.00 & $0.577 \pm 0.243$ & $7.35 \pm 10.83$ & $1.70 \pm 8.36$ & 40.32 \\

\ \ \ \ 192 
& 82.55 & $0.552 \pm 0.260$ & $7.59 \pm 11.21$ & 86.30 & $0.578 \pm 0.240$ & $7.58 \pm 11.14$ & $1.62 \pm 8.05$ & 45.47 \\

\midrule
Encoder initialization and adaptation & & & & & \\

\ \ \ \ Random initialization + frozen
& 58.73 & $0.365 \pm 0.299$ & $10.88 \pm 14.85$ & 65.64 & $0.409 \pm 0.286$ & $11.12 \pm 14.94$ & $5.47 \pm 15.14$ & 37.97 \\

\ \ \ \ Random initialization + training
& 70.91 & $0.449 \pm 0.287$ & $9.61 \pm 12.68$ & 76.12 & $0.482 \pm 0.269$ & $9.80 \pm 12.70$ & $3.99 \pm 13.65$ & 37.98 \\

\ \ \ \ Frozen pretrained
& 82.06 & $0.552 \pm 0.263$ & $7.27 \pm 10.56$ & 85.33 & $0.575 \pm 0.246$ & $7.36 \pm 10.69$ & $1.75 \pm 8.39$ & 37.85 \\

\ \ \ \ LoRA adaptation
& 82.61 & $0.553 \pm 0.260$ & $7.85 \pm 11.19$ & 85.21 & $0.572 \pm 0.244$ & $8.00 \pm 11.33$ & $2.11 \pm 9.34$ & 39.37 \\

\ \ \ \ Full fine-tuning
& \textbf{85.94} & $0.584 \pm 0.245$ & $5.81 \pm 9.42$ & \textbf{89.64} & $0.610 \pm 0.222$ & $5.73 \pm 9.28$ & $\textbf{0.84} \pm 5.18$ & 37.84 \\

\midrule

Input modalities & & & & & \\
\ \ \ \ RGB only
& 76.97 & $0.505 \pm 0.275$ & $8.28 \pm 12.15$ & 81.94 & $0.535 \pm 0.254$ & $8.28 \pm 12.01$ & $2.67 \pm 10.52$ & 16.76 \\

\ \ \ \ Depth only
& 76.73 & $0.495 \pm 0.278$ & $8.66 \pm 11.73$ & 81.58 & $0.525 \pm 0.258$ & $8.71 \pm 11.73$ & $2.84 \pm 11.49$ & 16.78 \\

\ \ \ \ Semantic only
& 30.06 & $0.186 \pm 0.265$ & $15.44 \pm 19.29$ & 34.48 & $0.215 \pm 0.271$ & $16.27 \pm 19.68$ & $9.15 \pm 19.80$ & 16.46 \\

\ \ \ \ RGB + depth
& 81.76 & $0.548 \pm 0.266$ & $6.84 \pm 10.17$ & 86.18 & $0.579 \pm 0.241$ & $6.93 \pm 10.29$ & $1.87 \pm 9.20$ & 26.36 \\

\ \ \ \ RGB + semantic
& 79.15 & $0.508 \pm 0.263$ & $8.76 \pm 12.01$ & 83.03 & $0.534 \pm 0.244$ & $8.81 \pm 11.88$ & $2.75 \pm 10.60$ & 26.26 \\

\ \ \ \ Depth + semantic
& 76.97 & $0.502 \pm 0.278$ & $9.03 \pm 11.59$ & 81.21 & $0.528 \pm 0.259$ & $9.06 \pm 11.58$ & $2.62 \pm 10.96$ & 26.31 \\

\bottomrule
\end{tabular*}
\caption{
Ablation studies on the design components of our approach.
Accuracy requires IoU $\geq 0.25$ and angular error $\leq 30^\circ$.
Angle statistics use positive-overlap matches.
Best mean results are bold.
}
\label{tab:ablations}
\end{table*}

\subsection{Baseline Comparison}
We compare LAGR against representative methods spanning four complementary approaches to robotic grasping. 
Task-agnostic grasp detection is represented by SE-ResUNet \cite{yu2022se}, which predicts grasp configurations based on visual and geometric cues without considering downstream task requirements. For a fair comparison under task-conditioned evaluation, we apply our language-conditioned reranker to SE-ResUNet's generated grasp proposals and report the resulting performance metrics.
Task-oriented grasping methods, including GATER \cite{sun2021gater}, OS-TOG \cite{holomjova2023one}, and ShapeGrasp \cite{li2024shapegrasp}, incorporate task semantics through object--action relationships, affordance transfer, and geometric decomposition, respectively. 
Language-conditioned grasping approaches, including CROG \cite{tziafas2023language}, LGD \cite{vuong2024language}, and RealVLG-R1 \cite{li2026realvlg}, leverage vision-language representations to predict grasp configurations from natural-language instructions. 
Finally, affordance-based approaches, including RAGNet \cite{wu2025ragnet}, GLOVER++ \cite{ma2025glover++}, and A4-Agent \cite{zhang2026a4}, emphasize identifying task-relevant interaction regions through semantic grounding, human demonstrations, and multimodal reasoning. 
For approaches that predict only the affordance region (either in mask or heatmap form), we couple them with a simple PCA based grasp estimator that outputs a grasp rectangle similar to other approaches.
Together, these baselines provide a comprehensive comparison across physical grasp detection, task-oriented reasoning, language-conditioned grasp prediction, and affordance localization, allowing us to evaluate the effectiveness of LAGR's explicit separation of physical feasibility and task-dependent grasp selection.

Table~\ref{tab:main_results} presents the quantitative comparison of LAGR against existing grasping approaches. LAGR achieves the highest task-conditioned success rate of 85.9\%, outperforming the strongest baseline, GATER (76.6\%), by 9.3 percentage points, while improving grasp IoU from 0.51 to 0.58. Under the grasp-stability evaluation, LAGR achieves 89.6\% success rate and 0.61 IoU, demonstrating consistent performance across both evaluation settings. Furthermore, LAGR achieves the lowest GCR (0.84\%), compared to 1.59\% for OS-TOG and 1.93\% for GATER, indicating reduced projected gripper--object interference.
LAGR also returns valid predictions for all 1,650 test samples, whereas OS-TOG fails to produce predictions for 632 samples. In terms of computational efficiency, LAGR requires 37.84 ms/sample, offering substantially lower inference latency than OS-TOG (3789.34 ms/sample), although SE-ResUNet and GATER remain faster at 9.15 and 23.18 ms/sample, respectively. Overall, these results demonstrate LAGR's effectiveness in task-conditioned grasp selection while maintaining high geometric alignment, low projected collision rates, and complete prediction coverage.

\subsection{Ablation Studies}
We investigate the contributions of our architectural components, encoder configurations, and input modalities towards LAGR's performance. 
Table~\ref{tab:ablations} reports our findings.

\noindent\textbf{Architectural Components}. 
We ablate key architectural components to assess their contributions to LAGR's performance. Table~\ref{tab:ablations} showcases our results. 
Selecting grasps solely by physical confidence reduces accuracy from 82.06\% to 53.03\%, highlighting the importance of task conditioning on grasp selection. 
Removing explicit proposal geometry and the proposal-context transformer decreases accuracy to 74.00\% and 77.76\% respectively, demonstrating the benefits of geometric representations and joint reasoning over grasp candidates. 
Excluding the physical-quality score results in a smaller reduction to 80.79\%, suggesting that visual and geometric features already capture substantial information about grasp feasibility.

\noindent\textbf{Reranker Loss Components}.
We investigate the contribution of individual reranker loss terms to task-conditioned grasp prediction (Table~\ref{tab:ablations}). 
Using calibration loss alone achieves 77.52\% accuracy, while incorporating listwise ranking improves performance to 82.24\%, highlighting the importance of learning relative preferences among grasp candidates. 
Adding a margin loss provides negligible improvement over calibration alone (77.94\%) and does not improve performance when combined with listwise ranking. 
Consequently, our final objective (\cref{eq:loss}) retains only calibration and listwise ranking losses, achieving a simpler formulation without sacrificing performance.

\noindent\textbf{Number of Grasp Proposals}.
We investigate the effect of varying the number of generated grasp proposals ($n$) on task-conditioned performance and computational efficiency (Table~\ref{tab:ablations}). 
Increasing $n$ from 1 to 64 improves accuracy from 53.03\% to 82.36\%, highlighting the importance of providing diverse candidates for task-conditioned selection. 
Beyond 64 proposals, performance largely saturates, with $n=128$ achieving the highest accuracy (82.61\%) and $n=192$ offering no further improvement (82.55\%). 
Meanwhile, inference time increases with the number of proposals, reaching 45.47 ms/sample at $n=192$. 
LAGR uses $n=96$ as the default configuration, providing a practical balance between accuracy, and computational cost.

\noindent\textbf{Encoder Initialization and Adaptation}.
We investigate the impact of multimodal pretraining and encoder adaptation on task-oriented grasp prediction (see Table~\ref{tab:ablations}). 
Full fine-tuning achieves the highest accuracy (85.94\%), outperforming the frozen pretrained encoder (82.06\%) of proposed approach, highlighting the benefits of adapting visual representations to grasp-specific geometry. 
LoRA adaptation yields only marginal improvement (82.61\%), suggesting that limited parameter updates provide less effective task specialization. 
Training from random initialization reduces accuracy to 70.91\%, with a further decline to 58.73\% when the randomly initialized encoder is frozen. 
These results underscore the importance of multimodal pretraining and the additional benefits of end-to-end adaptation.

\noindent\textbf{Input Modalities}.
We evaluate the contribution of individual visual modalities and their combinations to task-oriented grasp prediction (Table~\ref{tab:ablations}). 
Combining RGB and depth achieves 81.76\% accuracy, closely approaching the full RGB--depth--semantic configuration (82.06\%), while reducing inference time from 37.84 to 26.36 ms/sample. 
RGB-only (76.97\%) and depth-only (76.73\%) configurations perform worse, highlighting the complementary roles of appearance and geometric information. 
Semantic-only input yields substantially lower accuracy (30.06\%), indicating that semantic labels alone lack the fine-grained information required for precise grasp localization. 
Adding semantics to RGB or depth provides modest improvements, suggesting that semantic information offers limited additional benefit when visual and geometric features are available.

\subsection{Real-World Franka Research 3 Deployment}
We evaluate LAGR on a Franka Research 3 robotic arm equipped with a two-finger parallel-jaw gripper across five objects and ten object--task pairs, with ten trials per pair (Table~\ref{tab:metrics-rw}). 
LAGR achieves an overall grasp success rate of 93\%, outperforming GATER (72\%) by 21 percentage points. 
Notably, LAGR demonstrates substantial improvements in handover tasks, achieving 92\% success compared to 58\% for GATER. 
The largest improvements are observed for pan handover (100\% vs. 0\%) and jug handover (60\% vs. 0\%), highlighting LAGR's ability to select grasps appropriate for different manipulation objectives. 
Both methods achieve comparable performance on several functional tasks, including pouring and screwdriver operation, while LAGR achieves perfect success across eight of the ten object--task pairs. 
These results demonstrate the effectiveness of LAGR's task-conditioned grasp selection in real-world robotic manipulation, although the lower success rates on jug manipulation indicate remaining challenges.

\begin{table}[t]
    \centering
    \footnotesize
    \setlength{\tabcolsep}{8pt}
    \renewcommand{\arraystretch}{1.12}
    \resizebox{\columnwidth}{!}{%
    \begin{tabular}{@{}llcc@{}}
    \toprule
    \textbf{Object} & \textbf{Task}
    & \textbf{LAGR} & \textbf{GATER}\\
    \midrule
    \multirow{2}{*}{Hammer}
        & Pound & 10/10 & 8/10 \\
        & Handover & 10/10 & 9/10 \\
    \midrule
    \multirow{2}{*}{Jug}
        & Pour & 7/10 & 7/10 \\
        & Handover & 6/10 & 0/10 \\
    \midrule
    \multirow{2}{*}{Pan}
        & Cook & 10/10 & 9/10 \\
        & Handover & 10/10 & 0/10 \\
    \midrule
    \multirow{2}{*}{Screwdriver}
        & Tighten/loosen & 10/10 & 10/10 \\
        & Handover & 10/10 & 10/10 \\
    \midrule
    \multirow{2}{*}{Spoon}
        & Scoop & 10/10 & 9/10 \\
        & Handover & 10/10 & 10/10 \\
    \midrule
    \multicolumn{2}{l}{\textbf{Overall}}
        & \textbf{93/100} & 72/100 \\
    \bottomrule
    \end{tabular}%
    }
    \caption{Real-world task-oriented grasping performance
    using a Franka Research 3 with a parallel-jaw gripper.
    Each object is evaluated on two tasks, with ten
    orientations per task. Results indicate successful
    grasps out of ten trials per object--task pair.}
    \label{tab:metrics-rw}
    \vspace{-5mm}
\end{table}

\section{Conclusion}
\label{sec:conclusion}
We introduced LAGR, a task-oriented grasp generation framework that explicitly separates physical grasp feasibility from task-dependent suitability. 
To address the lack of dense task-conditioned supervision, we constructed a dataset comprising 2.68 million scored grasp annotations across 100,911 image–task pairs. 
Leveraging this dataset, LAGR combines multimodal grasp proposal generation with language-conditioned reranking to select geometrically plausible and functionally appropriate grasps. 
Extensive experiments demonstrate the effectiveness of LAGR, achieving 85.94\% task-conditioned grasp accuracy, with further validation through real-world robotic grasp execution. 
However, LAGR's performance depends on the quality and diversity of the generated grasp proposals, and its current formulation is limited to two-finger parallel-jaw grasping.
Future work will explore more adaptive and diverse grasp generation strategies and extend the framework to multi-fingered robotic hands for broader applicability.

{
    \small
    \bibliographystyle{ieeenat_fullname}
    \bibliography{main}
}

\clearpage

\end{document}